\documentclass{article}

\PassOptionsToPackage{sort, numbers, compress}{natbib}

 \usepackage[main, final]{neurips_2026}

\usepackage[utf8]{inputenc} 
\usepackage[T1]{fontenc}    
\usepackage{hyperref}       
\usepackage{url}            
\usepackage{booktabs}       
\usepackage{amsfonts}       
\usepackage{nicefrac}       
\usepackage{microtype}      
\usepackage{xcolor}         
\usepackage{seqsplit}

\usepackage{amsmath,amsfonts,bm}

\def\1{\bm{1}}

\DeclareMathAlphabet{\mathsfit}{\encodingdefault}{\sfdefault}{m}{sl}
\SetMathAlphabet{\mathsfit}{bold}{\encodingdefault}{\sfdefault}{bx}{n}

\newcommand{\cond}{\kappa}      

\usepackage{multirow}
\usepackage{graphicx}
\usepackage{subfigure}
\usepackage{amsmath}
\usepackage{amssymb}
\usepackage{mathtools}
\usepackage{amsthm}
\usepackage[capitalize,noabbrev]{cleveref}
\usepackage{fancyhdr}
\usepackage{algorithm}
\usepackage{eso-pic}
\usepackage{forloop}
\usepackage{url}
\usepackage{makecell}

\usepackage[normalem]{ulem}
\usepackage{stfloats}

\theoremstyle{plain}

\theoremstyle{definition}

\theoremstyle{remark}

\usepackage{algpseudocode}
\usepackage{caption}

\algrenewcommand\algorithmicrequire{\textbf{Input:}}
\algrenewcommand\algorithmicensure{\textbf{Output:}}

\usepackage[utf8]{inputenc}
\usepackage{newunicodechar}
\newunicodechar{−}{\textminus}

\newcommand{\name}{{$\cond$}-LoRA}
\newcommand{\namebf}{\bf {$\bm \cond$}-LoRA}
\newcommand{\fullLoRA}{{\small \textit{full}}-LoRA}

\title{\namebf:Condition Numbers Reveal \\Which LoRA Matrices Worth Updating}

\author{%
  Jianghui Wang$^1$\thanks{Equal contribution.}, Silong Yong$^{2*}$,
  Francesco Orabona$^1$, Marco Canini$^1$, Katia P. Sycara$^{2}$, Yaqi Xie$^{2}$\thanks{Corresponding author.} \vspace{5pt}\\
   $^1$King Abdullah University of Science and Technology \vspace{5pt}\\
   $^2$Carnegie Mellon University \vspace{5pt}\\
    \{jianghui.wang, francesco.orabona, marco\}@kaust.edu.sa\\
    \{silongy, sycara, yaqix\}@andrew.cmu.edu
}

\begin{document}

\maketitle

\begin{abstract}

Low-Rank Adaptation (LoRA) has become a widely adopted technique for efficient neural network fine-tuning, decomposing model updates into low-rank matrices. However, LoRA remains computationally costly because it updates all matrices uniformly, regardless of their actual contribution to adaptation. This cost is especially prohibitive for large-scale models with billions of parameters and for resource-constrained settings such as edge deployment and on-device fine-tuning. We show for the first time that not all LoRA matrices are equally worth tuning: matrices with smaller condition numbers (the ratio of largest to smallest singular value) are already well-balanced across directions and contribute only marginally to adaptation, whereas matrices with larger condition numbers contain underdeveloped directions that span richer subspaces and drive most of the performance gains. This observation itself is a key contribution of our work, and it motivates a more selective approach to fine-tuning.
Building on this insight, we propose {\namebf}, a method that optimizes LoRA by focusing updates on the matrices with the largest condition numbers, which capture the most informative directions of change. 
By restricting LoRA updates to the top 50\% of weight matrices ranked by condition number, {\name} halves the trainable parameter count and correspondingly reduces compute and memory cost. Extensive experiments across multiple benchmarks show that this design cuts fine-tuning time by 16.2\% on average while matching the accuracy of standard LoRA and reducing memory cost by 4.5\%. Further analysis reveals that the condition numbers of the selected matrices consistently decrease over training, suggesting that {\name}'s effectiveness stems from targeted spectral rebalancing rather than parameter selection alone.
\end{abstract}

\section{Introduction}\label{sec:intro}
\looseness=-1
Fine-tuning large-scale pre-trained models has become the cornerstone of state-of-the-art performance in Natural Language Processing, Computer Vision, and many other domains~\cite{brown2020language, devlin2019bert, dosovitskiy2020image}. As model sizes grow, however, the computational resources required for fine-tuning increase rapidly, creating significant challenges in both training time and memory usage~\cite{kaplan2020scaling}. Traditional fine-tuning approaches update all parameters of the model, incurring high computational and memory costs. Parameter-Efficient Fine-Tuning (PEFT) methods have emerged to alleviate this burden, including adapter modules~\cite{houlsby2019parameter}, prefix tuning~\cite{li2021prefix}, prompt tuning~\cite{lester2021power}, and Low-Rank Adaptation (LoRA)~\cite{hu2021lora}. While these techniques substantially reduce the number of trainable parameters, LoRA has emerged as one of the most widely adopted PEFT methods due to its simplicity, compatibility with Transformer architectures, and strong empirical performance across a broad range of downstream tasks. However, existing LoRA methods still leave considerable room for improvement in how adaptation capacity is allocated across the model.
\looseness=-1
In particular, LoRA approximates parameter updates with low-rank matrices, significantly reducing the number of trainable parameters adapted during fine-tuning~\cite{hu2021lora}. Despite its successes, LoRA introduces certain inefficiencies. In current practice, LoRA is applied uniformly across every layer of the model, extending to each projection in Transformer architectures~\cite{vaswani2017attention}. For instance, in the self-attention layers, LoRA is added to the query ($W_Q$), key ($W_K$), value ($W_V$), and output ($W_O$) projection matrices, where $W_Q$ and $W_K$ produce the query and key vectors that determine attention scores between tokens, $W_V$ projects token features into the values aggregated according to those scores, and $W_O$ recombines the multi-head outputs back into the residual stream. In the Multi-Layer Perceptron (MLP) layers, LoRA is applied to the up- and down-projection matrices, which expand and compress feature dimensions, as well as to the gating layers that modulate feature flow. As models grow larger, the number of LoRA modules required grows proportionally. This widespread application leads to an excessive proliferation of LoRA modules, causing unnecessary computational overhead, wasting resources, and potentially hindering convergence. By updating all low-rank matrices, including those with minimal impact on model performance, LoRA introduces redundant computations that slow down convergence, especially at scale. More importantly, different matrices exhibit highly uneven sensitivity to parameter updates, making uniform allocation of LoRA capacity inherently suboptimal.

To overcome these limitations, we propose {\name}, a method that builds on LoRA while addressing its inefficiencies. {\name} introduces a condition number-based selection mechanism that prioritizes matrices with greater impact on the model's behavior. Formally, the condition number of a matrix $W$ is defined as the ratio of its largest to its smallest singular value,
\begin{equation}
\kappa(W) \;=\; \frac{\sigma_{\max}(W)}{\sigma_{\min}(W)},
\end{equation}
and is a standard tool in numerical linear algebra for quantifying the sensitivity to perturbation~\cite{trefethen2022numerical}. Intuitively, $\kappa(W)$ measures how unevenly $W$ scales different directions in its input space. A well-conditioned matrix (small $\kappa$) scales all directions roughly uniformly and produces stable transformations. An ill-conditioned matrix (large $\kappa$) stretches some directions while severely compressing others, so that perturbations along certain directions are greatly amplified while information along other directions is suppressed. Crucially for our setting, ill-conditioned matrices admit directions along which small parameter updates can induce disproportionately large changes in the output, which makes them highly responsive to low-rank adaptation.

Building on this, we hypothesize that ill-conditioned matrices are the primary sources of inefficiency and instability during adaptation, and that allocating LoRA capacity to them yields the largest return per unit of compute. We support this hypothesis both theoretically, by analyzing how spectral properties bound the effect of a low-rank update on the matrix's action (Section~\ref{sec:setup_and_methodology}), and empirically, we observe that LoRA fine-tuning of selected high-$\kappa$ matrices reduces their condition numbers over the course of training, suggesting that the adaptation process actively rebalances the underlying transformations toward better-conditioned and more stable representations. The selection procedure itself is lightweight and requires no training-time signal: for each candidate projection matrix in the pre-trained model, we compute its singular values once before fine-tuning and derive $\kappa(W)$. We then rank matrices within a layer and apply LoRA only to the top fraction of matrices under a given adaptation budget, leaving the rest frozen. Because the criterion depends solely on the pre-trained weights, the selection step adds negligible overhead and can be reused across downstream tasks.

The contributions of our study are threefold. First, we identify a spectral asymmetry in pre-trained transformers: weight matrices within the same architectural group exhibit a wide range of condition numbers, and matrices with larger condition numbers admit directions where bounded low-rank updates can induce disproportionately large changes. Second, we introduce {\name}, a method that exploits this asymmetry by selectively adapting matrices ranked by condition number, supported by a closed-form sensitivity bound and a training-free, one-shot selection procedure. Third, we validate {\name} across multiple models and benchmarks, and further show that the condition numbers of adapted matrices decrease substantially over the course of fine-tuning, providing empirical evidence that our selection mechanism aligns with how LoRA reshapes the underlying transformations.

\section{Related work}\label{sec:related_works}

Parameter-Efficient Fine-Tuning (PEFT) addresses the cost of adapting large pre-trained models by training only a small subset of parameters. Common approaches include adapter modules~\citep{houlsby2019parameter}, prefix-tuning~\citep{li2021prefix}, and prompt-tuning~\citep{lester2021power}. Low-Rank Adaptation (LoRA)~\citep{hu2021lora} is the most widely used: it adds trainable low-rank matrices to existing linear layers, keeps pre-trained weights frozen, and incurs no inference overhead after merging.
A range of extensions modify LoRA along different axes. AdaLoRA~\citep{zhang2023adaptive} reallocates rank budget across layers based on importance scores. Delta-LoRA~\citep{zi2023delta} propagates adapter updates back into the original weights. LoSparse~\citep{li2023losparse} combines low-rank updates with structured pruning. DoRA~\citep{liu2024dora} decomposes weights into magnitude and direction and adapts only the direction. SoRA~\citep{ding2023sparse} gates the rank of each adapter during training. QLoRA~\citep{dettmers2024qlora} and QA-LoRA~\citep{xu2023qa} combine LoRA with quantization; {\name} is orthogonal to these and can be stacked on top.
Several recent methods use the spectrum of pre-trained weights to inform adaptation. PiSSA~\citep{meng2024pissa} initializes the low-rank matrices $A$ and $B$ from the principal singular vectors of $W$ to align updates with the dominant subspace. MiLoRA~\citep{wang2025milora} does the opposite, initializing from the minor components to preserve pre-trained knowledge. SVFT~\citep{lingam2024svft} parameterizes $\Delta W$ as a sparse combination of outer products of $W$'s singular vectors. These methods use spectral information to shape how LoRA is parameterized; they still adapt every candidate matrix.
A closer line of work asks which components are worth adapting at all. LISA~\citep{pan2024lisa} samples layers to update from a layerwise importance distribution. LoRA-Drop~\citep{zhou2025lora} prunes adapters whose output activations are small. AFLoRA~\citep{liu2024aflora} freezes low-rank paths during training using a gradient-and-uncertainty score. HiFi~\citep{gui2023hifi} ranks attention heads by PageRank importance, though without LoRA. AdaLoRA~\citep{zhang2023adaptive} can also be read as a selection method, since it allocates rank budget across modules. All of these signals come from training-time statistics or apply at the layer level. {\name} uses a structural property of the pre-trained weights, computed once offline, and operates at the level of individual projection matrices. To our knowledge, {\name} is the first to use the spectral conditioning of pre-trained weights as a selection signal for LoRA.
\section{Methodology}\label{sec:setup_and_methodology}

In this section, we formally introduce the necessary definitions and our method.

\subsection{Preliminary}
\label{sec:notation}

We summarize the notation used throughout the paper.

\paragraph{Matrices and norms} 
For a matrix $M \in \mathbb{R}^{m \times n}$, we denote its operator (spectral) norm by
\[
\|M\|_{\mathrm{2}} \; := \; \max_{\|x\|_2 = 1} \ \|Mx\|_2,
\]
which corresponds to the largest singular value of $M$. We denote the Frobenius norm by
\[
\|M\|_F \; := \; \sqrt{\sum_{i,j} M_{ij}^2},
\]
\looseness=-1
which measures the overall magnitude of matrix entries. The rank is denoted by $\mathrm{rank}(M)$. We write $I_d$ for the $d \times d$ identity matrix. Unless otherwise specified, $\|\cdot\|_2$ for vectors denotes the Euclidean norm.

\paragraph{Singular value decomposition} Every matrix $W \in \mathbb{R}^{m \times n}$ admits a singular value decomposition
\begin{equation*}
W = U \Sigma V^\top, \qquad \Sigma = \mathrm{diag}(\sigma_1, \ldots, \sigma_r),
\end{equation*}
\looseness=-1
with $\sigma_1 \geq \sigma_2 \geq \cdots \geq \sigma_r > 0$ and $r = \mathrm{rank}(W)$. The columns $u_i$ of $U \in \mathbb{R}^{m \times r}$ and $v_i$ of $V \in \mathbb{R}^{n \times r}$ are the left and right singular vectors, respectively. We write $\sigma_{\max}(W) = \sigma_1$ and $\sigma_{\min}(W) = \sigma_r$.

\paragraph{Spectral summaries} The condition number of $W$ is
\begin{equation*}
\kappa(W) \;:=\; \frac{\sigma_{\max}(W)}{\sigma_{\min}(W)}.
\end{equation*}
\looseness=-1
For a matrix $W$, let $\sigma_i(W)$ denote its singular values, and let $u_i, v_i$ denote the corresponding left and right singular vectors satisfying
\[
W v_i = \sigma_i u_i.
\]

\paragraph{LoRA parameterization} A LoRA adapter of rank $r$ on a base matrix $W \in \mathbb{R}^{m \times n}$ replaces $W$ during the forward pass with $W + \Delta W$, where
\begin{equation}
\Delta W = B A, \qquad B \in \mathbb{R}^{m \times r}, \quad A \in \mathbb{R}^{r \times n},
\end{equation}
and only $B, A$ are trained. Following the standard LoRA initialization~\cite{hu2021lora}, $B$ is initialized to zero while $A$ is randomly initialized, so that the initial update satisfies $\Delta W = 0$ and preserves the pretrained model at the start of the fine-tuning procedure. We refer to $r$ as the LoRA rank. $\|\Delta W\|_F$ constrains the magnitude of the update, which we refer to as the update's Frobenius budget.

\paragraph{Candidate set and selection} 
\looseness=-1
Let $\{W_1, \ldots, W_N\}$ denote the set of pretrained weight matrices eligible for LoRA adaptation, typically including attention projections and MLP linear layers across all transformer blocks. A selection is a subset $\mathcal{S} \subseteq \{1, \ldots, N\}$ with $|\mathcal{S}| = K$, where $K$ denotes the adaptation budget.

\begin{figure*}[t]
  \vskip 0.2in
  \begin{center}
    \includegraphics[width=\linewidth]{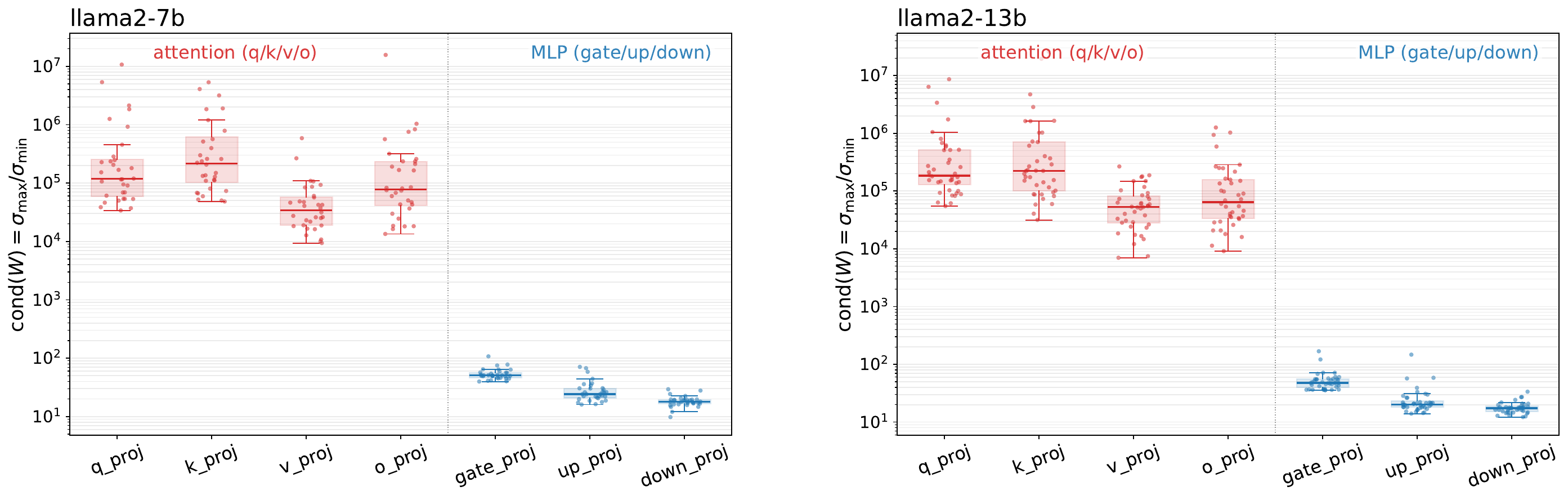}
    \caption{
    \looseness=-1
      Condition numbers of different projection matrices in transformer architectures exhibit clear clustering patterns across architectural components. Most importantly, attention projections (Query, Key, Value, and Output matrices) and MLP projections (up-, down-, and gate-projection matrices) form distinct condition number scales. This observation motivates our group-wise selection strategy, where matrices are ranked and selected within each architectural group rather than globally across the entire model.
    }
    \label{fig:cond}
  \end{center}
\end{figure*}

To ensure meaningful comparison across matrices, we partition the candidate set into architectural groups $\mathcal{G}_1, \ldots, \mathcal{G}_g$. In transformer architectures, we use separate groups for attention projections (e.g., query, key, value, and output matrices) and MLP projections (e.g., up-, down-, and gate-projection matrices). As illustrated in Figure~\ref{fig:cond}, matrices with different architectural roles exhibit distinct condition number distributions and naturally form separate clusters. We therefore perform ranking and selection within each group rather than globally across all matrices.

\paragraph{Reachable capacity} 
\looseness=-1
Given a matrix $W$ and a bounded low-rank update satisfying $\|\Delta W\|_F \le \epsilon$, we define $\rho(W;\epsilon)$ as the maximum relative change that can be induced in the action of $W$ along any singular direction. Intuitively, $\rho(W;\epsilon)$ measures how sensitive the behavior of $W$ is to a bounded low-rank perturbation.

\subsection{Method}
\label{sec:method}
\looseness=-1
Standard LoRA~\citep{hu2021lora} applies low-rank adapters uniformly to a predefined set of weight matrices (e.g., all attention and MLP projections), implicitly assuming that all matrices benefit equally from adaptation. We challenge this assumption by analyzing how much a bounded low-rank update can actually change the behavior of each matrix.

\paragraph{Effect of bounded low-rank updates}
Consider a pretrained weight matrix $W \in \mathbb{R}^{m \times n}$. Suppose we apply a bounded low-rank update $\Delta W$ such that $\mathrm{rank}(\Delta W) \le k$ and $\|\Delta W\|_F \le \epsilon$. We analyze how much this update can change the action of $W$ along a singular direction $v_i$:
\begin{equation*}
\frac{\|(W + \Delta W)v_i - W v_i\|}{\|W v_i\|}.
\end{equation*}

The numerator satisfies
\[
\|(W+\Delta W)v_i - Wv_i\|
=
\|\Delta W v_i\|
\le
\|\Delta W\|_2 \|v_i\|_2
=
\|\Delta W\|_2
\le
\|\Delta W\|_F
\le
\epsilon,
\]
where we use the fact that $v_i$ is a unit singular vector. The denominator is equal to
\[
\|Wv_i\|
=
\|\sigma_i u_i\|
=
\sigma_i.
\]
Therefore, the relative change along direction $v_i$ is upper bounded by $\frac{\epsilon}{\sigma_i}$.

\paragraph{Intuition}
This bound reveals a key asymmetry: directions corresponding to small singular values ($\sigma_i$) can be significantly altered even by small updates. In particular, the smallest singular value $\sigma_{\min}(W)$ determines the maximum achievable relative change:
\begin{equation*}
\rho(W; \epsilon) \; := \; \max_i \frac{\epsilon}{\sigma_i} = \frac{\epsilon}{\sigma_{\min}(W)}.
\end{equation*}

This quantity measures the maximum relative change that a bounded low-rank update can induce in the behavior of $W$: how much its functional behavior can be changed under a bounded low-rank update. Importantly, this upper bound is attainable by low-rank updates. In particular, a rank-1 update of the form $\Delta W = \epsilon\, u\, v_i^\top$ achieves the bound along direction $v_i$. Since LoRA parameterizes updates as low-rank matrices, it can realize such updates in practice. Therefore, $\rho(W;\epsilon)$ is not merely a theoretical upper bound, but reflects the actual expressive capacity of LoRA updates.

\looseness=-1
The maximum achievable relative change under a bounded low-rank update can therefore be written as
\begin{equation}
\rho(W; \epsilon)
\;:=\;
\max_i \frac{\epsilon}{\sigma_i}
\;=\;
\frac{\epsilon}{\sigma_{\min}(W)}
\;=\;
\frac{\epsilon \cdot \kappa(W)}{\sigma_{\max}(W)}.
\label{eq:reach}
\end{equation}

\paragraph{Connection to condition number}
Equation~\eqref{eq:reach} shows that the maximum achievable relative change under a bounded low-rank update is inversely proportional to the smallest singular value, and can be expressed in terms of the condition number
\[
\kappa(W)
= \frac{\sigma_{\max}(W)}{\sigma_{\min}(W)}.
\]
\looseness=-1
Under the mild assumption that $\sigma_{\max}(W)$ varies within a comparatively smaller range inside an architectural group, Equation~\eqref{eq:reach} suggests that the sensitivity of a matrix to bounded low-rank updates is largely governed by its condition number. Empirically, as illustrated in Figure~\ref{fig:cond}, matrices within the same architectural group exhibit more coherent condition number distributions compared to cross-group comparisons, making condition number a practical proxy for ranking adaptation sensitivity. This observation motivates our strategy of prioritizing matrices with larger condition numbers for LoRA adaptation.

Beyond the connection in Eq.~\eqref{eq:reach}, the condition number has another useful property: it is invariant to uniform scaling of the matrix. Specifically, for any scalar $c>0$,
\[
\kappa(cW)=\kappa(W).
\]
\looseness=-1
This property is desirable for matrix selection because the same network function can often be represented using different parameter scalings. For example, two consecutive linear transformations $W_2W_1$ can be equivalently rewritten as $(cW_2)(W_1/c)$ without changing the overall mapping~\citep{dinh2017sharp}. Under such rescaling, quantities based solely on $\sigma_{\min}(W)$ vary with the scale factor, potentially altering the ranking of matrices despite preserving the network function. In contrast, the condition number remains unchanged, making it a more intrinsic criterion for ranking matrices for LoRA adaptation.

\paragraph{Selection rule}
The above analysis suggests a simple principle: \emph{matrices with larger condition numbers admit directions where bounded low-rank updates can induce larger relative changes}. We therefore prioritize such matrices for LoRA adaptation.

\looseness=-1
Under a fixed budget of $K$ LoRA modules, our goal is to allocate adaptation updates to matrices that are most sensitive to bounded low-rank perturbations. To formalize this intuition, given a subset $\mathcal{S}$ of matrices, we define
\begin{equation*}
\Phi(\mathcal{S}) = \sum_{W_i \in \mathcal{S}} \rho(W_i; \epsilon),
\end{equation*}
which measures the total achievable relative change across the selected matrices.

Selecting matrices that maximize $\Phi(\mathcal{S})$ under the constraint $|\mathcal{S}| = K$ naturally prioritizes matrices with larger values of $\rho(W_i;\epsilon)$. Under the mild assumption that $\sigma_{\max}(W)$ varies within a comparatively smaller range inside each architectural group, this ranking is approximately consistent with selecting matrices with larger condition numbers.

\subsection{Algorithm}
\label{sec:algorithm}

\looseness=-1
Our procedure takes a pretrained model and a target selection ratio 
$\alpha \in (0,1]$, and outputs a fine-tuned model in which LoRA adapters are attached only to a selected subset of pretrained weight matrices. We first enumerate candidate weight matrices, such as attention projections and MLP projections. For each candidate matrix $W$, we compute its condition number $\kappa(W)$ once before fine-tuning. Then, we rank the matrices within each architectural group and select the top $\alpha$ fraction with the largest condition numbers. LoRA adapters are added only to these selected matrices, while all other pretrained weights remain frozen.

The selection step is data-free and training-free: it depends only on the spectral properties of the pretrained weights and does not require gradient statistics, calibration data, or additional forward passes. Since condition numbers are computed once offline, the additional SVD cost is negligible compared to the cost of fine-tuning large models. Algorithm~\ref{alg:kappa-lora} summarizes the full procedure.

\paragraph{Group-wise selection}
In transformer architectures, different types of weight matrices can exhibit substantially different spectral norm and condition number distributions due to differences in their functional roles. For example, attention projections and MLP projections often operate at different scales, making direct global ranking potentially biased toward certain architectural components.

To address this, we perform selection within each architectural group. Specifically, we partition matrices into architectural groups, such as attention projections (query, key, value, and output matrices) and MLP projections (up-, down-, and gate-projection matrices), compute condition numbers within each group, and select a fixed ratio $\alpha_j$ per group.

\looseness=-1
This design ensures that adaptation capacity is distributed across different functional components of the model, while still prioritizing matrices that are more sensitive to bounded low-rank updates within each group.

\paragraph{Discussion}
\looseness=-1
Intuitively, ill-conditioned matrices contain highly anisotropic transformations, where certain directions are under-utilized (small $\sigma_i$) and thus easier to modify. In contrast, well-conditioned matrices already behave more uniformly and are less sensitive to low-rank updates. While the condition number alone does not fully determine the importance of a matrix, it provides a simple and effective proxy for identifying where adaptation can have the largest impact. We empirically validate that this selection leads to improved efficiency and faster convergence. This also explains why prioritizing well-conditioned matrices (with small condition numbers) is less effective: their transformations are already stable and isotropic, leaving limited room for impactful low-rank adaptation.

\begin{algorithm}[t]
\caption{{\name} Algorithm}
\label{alg:kappa-lora}
\begin{algorithmic}[1]
\Require Pretrained model with candidate matrices $\{W_1, \ldots, W_N\}$ partitioned into architectural groups $\{\mathcal{G}_1, \ldots, \mathcal{G}_g\}$ (e.g., attention and MLP); per-group selection ratios $\{\alpha_j\}_{j=1}^{g}$ with $\alpha_j \in (0, 1]$; fine-tuning dataset $\mathcal{D}$
\Ensure Fine-tuned model with LoRA adapters on the selected subset
\For{$i = 1$ \textbf{to} $N$}
    \State Compute $\sigma_{\max}(W_i)$ and $\sigma_{\min}(W_i)$ via truncated SVD
    \State $\kappa_i \gets \sigma_{\max}(W_i) \,/\, \sigma_{\min}(W_i)$
\EndFor
\State $\mathcal{S} \gets \emptyset$
\For{$j = 1$ \textbf{to} $g$}
    \State $K_j \gets \lceil \alpha_j \, |\mathcal{G}_j| \rceil$
    \State $\mathcal{S}_j \gets$ indices of the $K_j$ matrices in $\mathcal{G}_j$ with the largest $\kappa_i$
    \State $\mathcal{S} \gets \mathcal{S} \cup \mathcal{S}_j$
\EndFor
\For{$i \in \mathcal{S}$}
    \State Initialize $B_i \in \mathbb{R}^{m_i \times r}$ as zero, $A_i \in \mathbb{R}^{r \times n_i}$ via Kaiming uniform
    \State Replace forward pass of $W_i$ with $W_i + B_i A_i$
\EndFor
\For{$i \notin \mathcal{S}$}
    \State Freeze $W_i$; no adapter attached
\EndFor
\State Train $\{B_i, A_i\}_{i \in \mathcal{S}}$ on $\mathcal{D}$ with the standard LoRA optimization recipe; keep all $W_i$ frozen
\State \Return Fine-tuned model
\end{algorithmic}
\end{algorithm}

\section{Experiments}\label{sec:exp}

\begin{table}[t]
\centering
\caption{{\name} on NLG tasks. The results are averaged over three runs, with standard deviations included. The GSM8K and MATH datasets share a math fine-tuned model, while HumanEval and MBPP use a code fine-tuned model. MT-Bench utilizes a conversation fine-tuned model. The "Avg. time" represents the average fine-tuning time for the three models.}
\label{table:NLP_results}
\resizebox{1.0\linewidth}{!}{
\begin{tabular}{cccccccc}
\toprule
\multirow{2}{*}{\textbf{Model}} & \multirow{2}{*}{\textbf{Method}} & \multirow{2}{*}{\textbf{Avg. Time(↓)}} & \multicolumn{2}{c}{\textbf{Math}} & \multicolumn{2}{c}{\textbf{Code}} & \multicolumn{1}{c}{\textbf{Conversation}} \\
\cmidrule(lr){4-5} \cmidrule(lr){6-7} \cmidrule(lr){8-8}
 & & & \textbf{GSM8K(↑)} & \textbf{MATH(↑)} & \textbf{HumanEval(↑)} & \textbf{MBPP(↑)} & \textbf{MT-Bench(↑)} \\
\midrule
\multirow{2}{*}{\texttt{LLaMA 2-7B}}
            & {\fullLoRA} & 7071s & \textbf{41.40±0.21} & 5.06±0.17 & \textbf{21.8±0.2} & 33.4±0.3 & 4.99±0.02 \\
            & {\name}   & \textbf{5948s(-15.9\%)} & 39.29±0.35 & \textbf{5.30±0.24} & 21.3±0.2 & \textbf{37.6±0.4} & \textbf{5.10±0.03} \\ 
\midrule
\multirow{2}{*}{\texttt{Mistral-7B}}
            & {\fullLoRA} & 7251s & 70.51±0.25 & 19.76±0.32 & \textbf{48.2±0.1} & 63.2±0.2 & 6.29±0.05 \\
            & {\name}    & \textbf{6119s(-15.6\%)} & \textbf{70.60±0.18} & \textbf{19.84±0.54} & 45.1±0.2 & \textbf{63.3±0.4} & \textbf{6.44±0.04} \\ 
\midrule
\multirow{2}{*}{\texttt{Gemma-7B}}
            & {\fullLoRA} & 7962s & \textbf{77.41±0.45} & 29.46±0.48 & \textbf{52.9±0.3} & 53.0±0.3 & 5.42±0.04 \\
            & {\name}   & \textbf{6603s(-17.0\%)} & 76.27±0.32 & \textbf{29.50±0.36} & 41.5±0.3 & \textbf{63.0±0.2} & \textbf{5.44±0.04} \\
\bottomrule
\end{tabular}
}
\end{table}

To demonstrate the broad effectiveness of the {\name} Algorithm, we conducted extensive experiments across Natural Language Generation (NLG) tasks and Natural Language Understanding (NLU) tasks. All experiments were conducted on NVIDIA GH200 GPUs (96 GB HBM). All LoRA have the scale weight $1$. Unless otherwise stated, {\name} has selection ratio $\alpha=0.5$. 

\subsection{NLG benchmarks}

\paragraph{Experimental setup}
We adopt the Alpaca~\cite{alpaca} implementation strategy and fine-tune with the AdamW optimizer~\cite{loshchilov2018decoupled} using a global batch size of 128, a learning rate of 2e-5, a cosine annealing schedule~\cite{LoshchilovH17,yu2023bag}, a warmup ratio of 0.03, and no weight decay, with a maximum sequence length of 512. LoRA modules use rank $r=128$.

\paragraph{Tasks and evaluation}
We evaluate {\name} algorithms across a diverse set of NLG tasks, see Table~\ref{table:NLP_results}. Specifically, we fine-tune LLaMA 2-7B~\citep{touvron2023llama}, Mistral-7B~\citep{jiang2023mistral}, and Gemma-7B~\citep{team2024gemma} on the MetaMathQA dataset~\citep{yu2023metamath} to benchmark their mathematical reasoning capabilities on GSM8K~\citep{cobbe2021gsm8k} and MATH~\citep{hendrycks2021measuring}. To assess coding proficiency, we further fine-tune these models on CodeFeedback~\citep{zheng2024opencodeinterpreter} and evaluate on HumanEval~\citep{chen2021evaluating} and MBPP~\citep{austin2021program}. For conversational ability, we train on WizardLM-Evol-Instruct~\citep{xu2023wizardlm} and evaluate on MT-Bench~\citep{zheng2024judging}. Following \citet{meng2024pissa}, all models are fine-tuned for a single epoch on a 100K-example subset of each dataset to bound training compute and ensure a fair comparison across methods. Due to the small size of HumanEval and MBPP, and to maintain consistency with prior work, we report all code tasks results with one decimal place.

\paragraph{Efficiency} As shown in Table~\ref{table:NLP_results}, {\name} reduces fine-tuning time by 15.6–17.0\% across all three backbones. This speedup follows directly from {\name}'s parameter footprint: it introduces only half as many trainable parameters as {\fullLoRA}, effectively halving the size of the adapted module. The reduction is stable across model families and grows mildly with model complexity (largest on Gemma-7B), indicating that the speedup stems from the method itself rather than backbone-specific interactions. While {\name} consistently improves training efficiency, we further examine whether these savings come at the cost of downstream performance.

\looseness=-1
\paragraph{Effectiveness} {\name} matches or surpasses {\fullLoRA} on 11 of 15 (model, task) pairs despite adapting only half of the LoRA modules. It attains the best MATH score on every backbone (+0.24, +0.08, +0.04) and the best MT-Bench score on every backbone (+0.11, +0.15, +0.02). This consistent improvement on MATH and MT-Bench suggests that adaptation sensitivity is concentrated in a relatively small subset of matrices, and that selectively adapting high-condition-number matrices can preserve and sometimes improve complex reasoning and instruction-following behaviors. The largest improvements appear on MBPP (+4.2 on LLaMA 2-7B and +10.0 on Gemma-7B), where broader and more diverse program synthesis targets may benefit from concentrating updates on the most sensitive matrices. Differences on GSM8K remain within 1.2 points of {\fullLoRA} and overlap within standard deviations on Mistral-7B. 
In contrast, gains on HumanEval are less consistent, with performance drops of 0.5--11.4 points depending on the backbone. This suggests that exact functional code synthesis may be less tolerant to sparse adaptation than broader coding benchmarks such as MBPP.
Despite task-specific variation, the overall efficiency-quality tradeoff remains consistent across three backbones with markedly different pre-training corpora and tokenizers. {\name} achieves an average 16.2\% reduction in fine-tuning time without measurable degradation in aggregate downstream quality, suggesting that the observed behavior reflects a general property of LoRA adaptation rather than backbone-specific effects.

\vspace{-5pt}

\subsection{NLU benchmarks}

\looseness=-1
We evaluate {\name}'s NLU capability on the GLUE benchmark~\cite{wang2018glue} with 
DeBERTa-v3-base~\cite{he2021debertav3}. The LoRA rank is $r=8$, the standard choice for DeBERTa on GLUE. Since we do not normalize for training set size, we report the total fine-tuning time aggregated over all eight tasks. The number of training epochs is set per-task: 20 for RTE, 10 for CoLA, MRPC, and STS-B, 3 for SST-2 and QNLI, and 2 for MNLI and QQP.

\looseness=-1
\paragraph{Efficiency} As shown in Table~\ref{tab:glue_results}, {\name} reduces the cumulative wall-clock time across the eight GLUE tasks from 4979s to 3872s, corresponding to a $22.2\%$ reduction under identical training settings. The magnitude exceeds the $15.6$--$17.0\%$ range observed on the three 7B NLG backbones as shown in  Table~\ref{table:NLP_results}. 

\paragraph{Effectiveness} 
As shown in Table~\ref{tab:glue_results}, {\name} matches {\fullLoRA} on GLUE while training only half of the LoRA projections. Aggregating the per-task numbers into a single GLUE-style score, {\fullLoRA} reaches $88.44\%$ and {\name} reaches $87.91\%$; the $0.53$-point gap collapses to $0.04$ points ($88.44\%$ vs.\ $88.40\%$) once the $277$-example RTE split, which is well known to be high-variance, is set aside. This small gap suggests that adaptation sensitivity is highly concentrated in a relatively small subset of matrices even for language understanding tasks.

\looseness=-1
On a per-metric basis, {\name} is the top method on $7$ of the $12$ reported numbers, with the most pronounced gains on CoLA, STS-B, and QQP, and stays within typical run-to-run variance on the remainder. These gains suggest that selectively adapting high-condition-number matrices is sufficient to preserve fine-grained linguistic and semantic representations across a broad range of understanding tasks. Overall, these results suggest that sparse adaptation guided by condition number remains highly effective for language understanding tasks, indicating that a substantial fraction of LoRA modules can be removed without measurable degradation in aggregate benchmark quality while reducing wall-clock cost by $22.2\%$.

\begin{table}[t]
\centering
\caption{{\name} on NLU tasks. Matthews correlation on CoLA, Pearson--Spearman mean on STS-B, accuracy--F1 mean on MRPC and QQP, and matched--mismatched mean on MNLI.}
\label{tab:glue_results}
\resizebox{\textwidth}{!}{%
\begin{tabular}{lcccccccccc}
\toprule
\multirow{2}{*}{\textbf{Method}} & \multirow{2}{*}{\textbf{Time(↓)}} & \textbf{CoLA} & \textbf{SST-2} & \textbf{MRPC} & \textbf{STS-B} & \textbf{RTE} & \textbf{QNLI} & \textbf{MNLI} & \textbf{QQP} \\
                &               & \textbf{(Matt↑)} & \textbf{(acc↑)} & \textbf{(acc↑ / F1↑)} & \textbf{(P↑ / S↑)} & \textbf{(acc↑)} & \textbf{(acc↑)} & \textbf{(m↑ / mm↑)} & \textbf{(acc↑ / F1↑)} \\
\midrule
{\fullLoRA}          & 4979s & 66.24 & \textbf{95.64} & \textbf{90.69 / 93.31} & 91.60 / 91.49 & \textbf{88.45} & 94.09 & 90.24 / \textbf{90.40} & 90.74 / 87.74 \\
{\name}  & \textbf{3872s(-22.2\%)} & \textbf{66.56} & 95.41 & 89.95 / 92.77 & \textbf{91.77 / 91.53} & 84.48 & \textbf{94.11} & \textbf{90.27} / 90.11 & \textbf{90.95 / 88.13} \\
\bottomrule
\end{tabular}%
}
\end{table}

\begin{figure}[t]
\centering
\begin{minipage}[t]{0.48\textwidth}
\vspace{-38mm}
  \centering
  \small
  \captionof{table}{
  \looseness=-1 Trainable parameter counts.  
  Projections is the number of LoRA projections; Trainable sums the LoRA $A,B$ parameters across those modules.}
  \label{tab:lora-params}
  \resizebox{\linewidth}{!}{%
  \begin{tabular}{cccccc}
    \toprule
    \textbf{Model}       & \textbf{Method}                      & \textbf{Proj.} & \textbf{Trainable} & \textbf{Total}    & \textbf{Train.\,\%} \\
    \midrule
    \multirow{2}{*}{\texttt{LLaMA 2-7B}}   & {\fullLoRA}        & 224     & 319.8\,M  & 7.06\,B  & 4.75\%     \\
       & {\name} & 112     & 159.9\,M  & 6.90\,B  & 2.37\%     \\
    \midrule
    \multirow{2}{*}{\texttt{Mistral-7B}}  & {\fullLoRA}        & 224     & 335.5\,M  & 7.58\,B  & 4.63\%     \\
      & {\name} & 112     & 180.4\,M  & 7.42\,B  & 2.49\%     \\
    \midrule
    \multirow{2}{*}{\texttt{Gemma-7B}}    & {\fullLoRA}         & 196     & 400.0\,M  & 8.94\,B  & 4.69\%     \\
      & {\name} &  98     & 200.0\,M  & 8.74\,B  & 2.34\%     \\
    \midrule
    \multirow{2}{*}{\texttt{DeBERTa-v3}} & {\fullLoRA}     &  72     &   1.33\,M  & 185.2\,M & 0.72\%     \\
      & {\name} &  36     &   0.66\,M  & 184.5\,M & 0.36\%     \\
    \bottomrule
  \end{tabular}
  }
\end{minipage}
\hfill
\begin{minipage}[t]{0.48\textwidth}
  \centering
  \includegraphics[width=\linewidth]{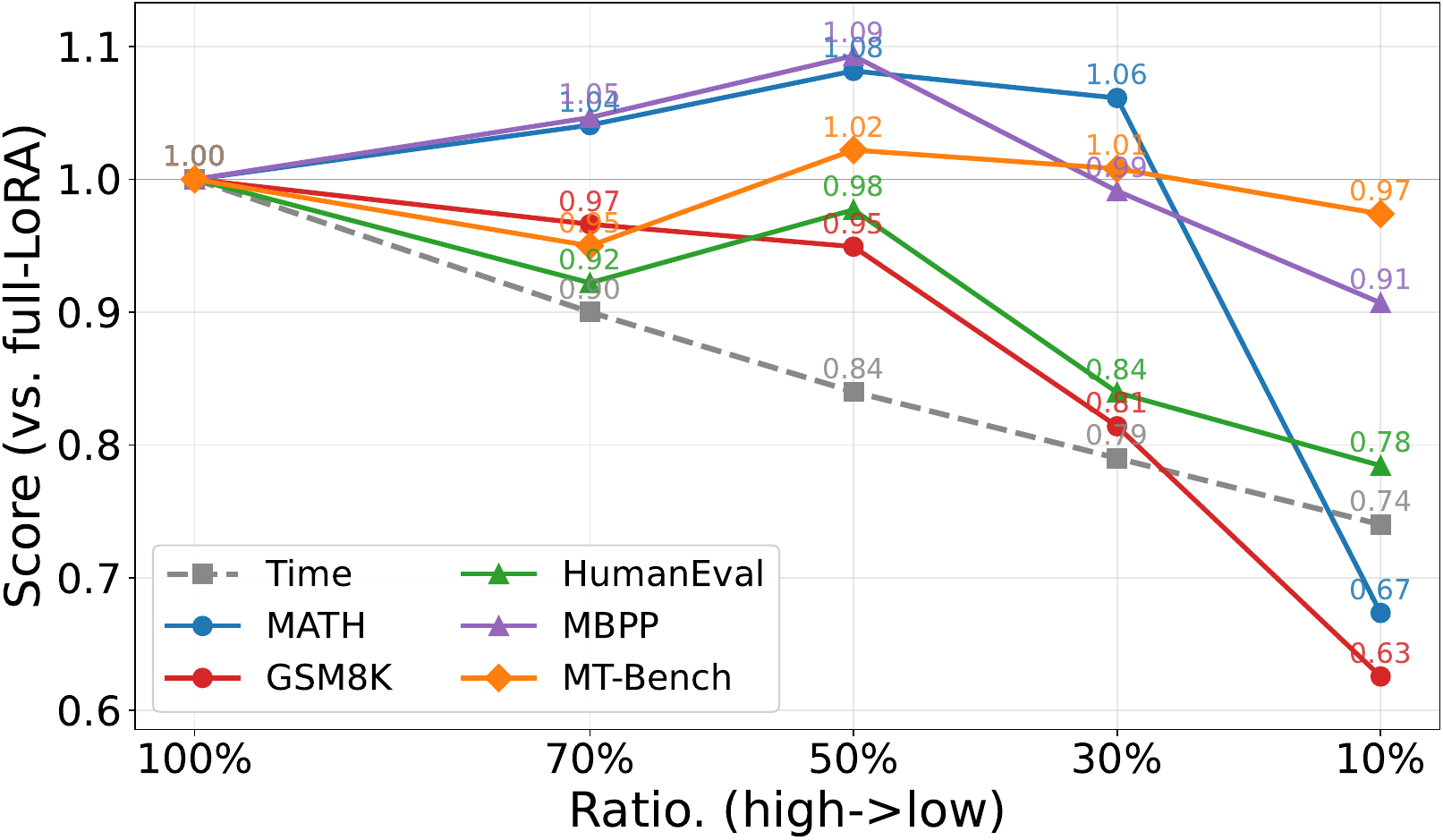}
  \captionof{figure}{{\name} sweep, normalized to full-LoRA at LLaMA2-7B.}
  \label{fig:ratio-sweep}
\end{minipage}

\end{figure}

\vspace{-5pt}
\subsection{Ablation Study}

\looseness=-1
We sweep the selection ratio $\alpha$ on LLaMA2-7B in Figure~\ref{fig:ratio-sweep}. $\alpha = 50\%$ lies near the Pareto knee, reducing wall-clock time by approximately $16\%$ while matching or exceeding {\fullLoRA} on four of five benchmarks.

\looseness=-1
When $\alpha$ becomes too small, performance degradation appears most noticeably on reasoning-intensive tasks such as MATH, suggesting that overly aggressive sparsification removes matrices that still contribute meaningfully to complex reasoning behaviors. In contrast, increasing $\alpha$ beyond $50\%$ yields diminishing performance gains while steadily increasing fine-tuning cost, indicating that many lower-priority matrices contribute only marginally once the highest-condition-number matrices are retained.

Overall, these results support the central hypothesis of {\name}: adaptation sensitivity is highly uneven across matrices, and a relatively small subset of high-condition-number matrices captures most of the useful adaptation capacity.

\vspace{-8pt}
\section{Discussion}\label{sec:conclusion}
\begin{figure}[t]
  \centering
  \includegraphics[width=\linewidth]{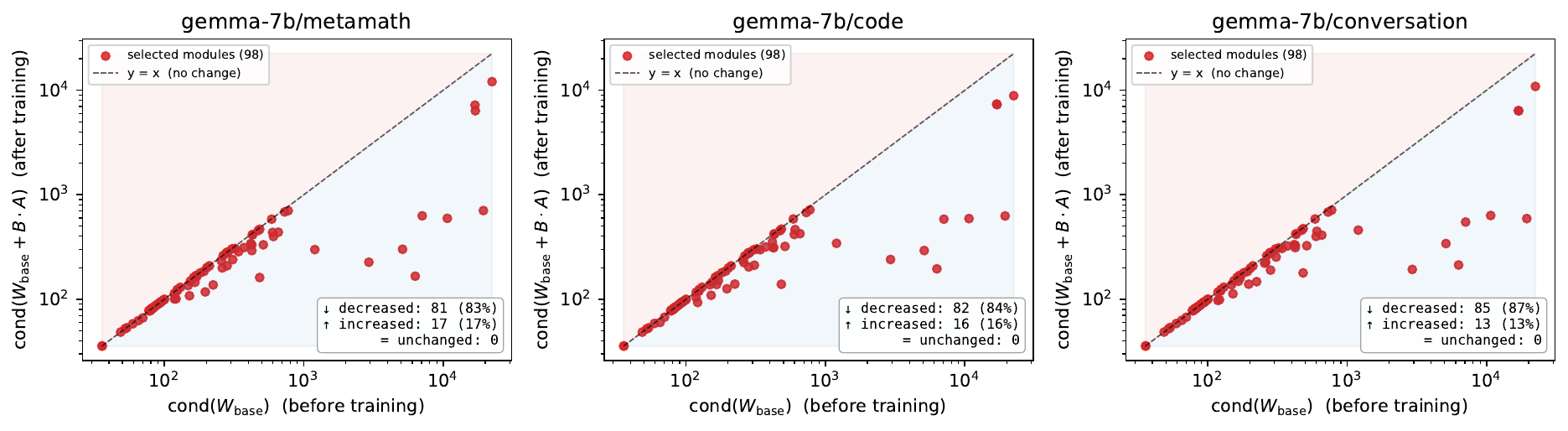}
  \caption{LoRA training systematically reduces the condition number of high-$\kappa$ projections in Gemma-7B. }
  \label{fig:cond-change}
  \vspace{-15pt}
\end{figure}

\vspace{-7pt}
A central design choice of {\name} (Section~\ref{sec:setup_and_methodology}) is to apply LoRA to the most ill-conditioned weight matrices, under the hypothesis that the low-rank updates help reduce their condition numbers. To validate this, we measure the change in condition number of all selected matrices in Gemma-7B before and after fine-tuning. As shown in Figure~\ref{fig:cond-change}, 83\%–87\% of these matrices become better conditioned, confirming our expectation and suggesting that the gains of {\name} stem from targeted spectral rebalancing rather than parameter selection alone.
Building on this insight, {\name} offers a simple recipe for efficient LoRA fine-tuning: rank weight matrices by condition number and adapt only the most ill-conditioned half. Across math, code, conversation, and NLU benchmarks, this design reduces fine-tuning time by 16.2\% on average while matching the accuracy of standard LoRA. We view condition-number–guided selection as a complementary axis to existing efficiency techniques, such as quantization, and leave their combination to future work.

\textbf{Limitations}\label{sec:limitation}
Our study has several limitations that open directions for future work. First, our analysis assumes that the condition number of pretrained weights remains a stable proxy for adaptation leverage throughout fine-tuning; a dynamic criterion that re-evaluates spectral statistics during training could, in principle, adapt better to objectives that significantly reshape weight geometry, such as long-horizon reinforcement learning. Second, we apply a fixed selection ratio $\alpha$ across all transformer blocks; allowing $\alpha$ to vary by depth, or combining {\name}'s matrix-level selection with rank-level allocation methods such as AdaLoRA, which are orthogonal to our criterion, may yield further gains. Finally, our theoretical analysis establishes that high-$\kappa$ matrices admit provably greater leverage under bounded low-rank perturbations rather than full optimality of top-$\kappa$ selection in the general case, and a formal characterization of when the criterion is provably optimal remains open.

\clearpage

\bibliographystyle{plainnat}
\bibliography{references}

\clearpage
\appendix

\begin{appendix}
	
	\thispagestyle{plain}
	\begin{center}
		{\Large \bf Appendix}
	\end{center}
	
\end{appendix}

\section{Memory and compute requirements}

We report our experiment memory usage at Table~\ref{tab:disk}. Base (MB) is the canonical
  Hugging Face checkpoint (\texttt{*.safetensors} only; TinyLlama is
  released in fp32 so its base is comparatively large for its parameter
  count). $\Delta$ = {\fullLoRA} - {\name} is
  the per-cell adapter saving in MB; $\Delta / \texttt{full}$ shows that the
  Mistral and TinyLlama ratios deviate slightly from $0.50$ because
  grouped-query attention (Mistral) and the smaller hidden size of TinyLlama
  perturb the per-module parameter mass. The final column expresses the
  saving as a fraction of the base-model storage cost: switching from
  {\fullLoRA} to {\name} reclaims roughly
  $4$--$5\%$ of base-model disk per cell, and the saving compounds across our method $\times$ backbone $\times$ task $\times$ rank sweep.

\begin{table}[ht]
  \centering
\caption{
  On-disk adapter checkpoint sizes at $r{=}128$ in fp32 (the PEFT
  default), alongside the published bf16/fp16 base-model size that has to be
  shipped to deploy each adapter.
  }
  \resizebox{\textwidth}{!}{%
  \begin{tabular}{lrrrrrr}
    \toprule
    \textbf{Backbone}        & \textbf{Base (MB)} & \textbf{{\fullLoRA} (MB)} & \textbf{{\name} (MB)} & \textbf{$\Delta$ (MB)} & \textbf{$\Delta$ / \texttt{full}} & \textbf{$\Delta$ ratio} \\
    \midrule
    TinyLlama-1.1B  &  4\,197 &  386 &  211 &  175 & $0.45$ & $4.2\%$ \\
    Llama-2-7B      & 12\,854 & 1\,221 &  611 &  610 & $0.50$ & $4.7\%$ \\
    Mistral-7B & 13\,814 & 1\,284 &  692 &  592 & $0.46$ & $4.3\%$ \\
    Gemma-7B        & 16\,288 & 1\,563 &  800 &  763 & $0.49$ & $4.7\%$ \\
    Llama-2-13B     & 24\,827 & 1\,911 &  956 &  955 & $0.50$ & $3.8\%$ \\
    \bottomrule

  \end{tabular}
  }

  \label{tab:disk}
\end{table}

All training and evaluation in this paper are run on a single NVIDIA
GH200 480\,GB Grace Hopper Superchip node; the relevant specifications are
summarised in Table~\ref{tab:hardware}. The ``480\,GB'' in the SKU
designation refers to the Grace LPDDR5X host memory; the GPU side of the
chip exposes $96$\,GB of HBM3, which is the figure that actually bounds our
training workloads. We use bf16 mixed precision with Flash-Attention~2 and
gradient checkpointing enabled throughout, which leaves substantial
headroom on the $96$\,GB HBM3 (peak GPU usage we observe is $\approx
21$\,GB at $7$B and $\approx 39$\,GB at $13$B). Inference for the
generation evaluations runs on the same node via vLLM.

\begin{table}[ht]
  \centering
  \caption{Hardware platform. The 96\,GB HBM3 capacity is the ceiling that
  actually bounds our training workloads; the much larger $480$\,GB Grace
  memory is reachable from the GPU via the $900$\,GB/s NVLink-C2C link but
  is not on the critical path for the experiments reported here. The GPU
  die is the same Hopper SXM die used in NVIDIA H100, so per-GPU
  throughput is comparable to a single H100 SXM card.}
  \begin{tabular}{ll}
    \toprule
    \textbf{Component}                       & \textbf{Specification} \\
    \midrule
    GPU                             & NVIDIA GH200 480\,GB (Grace Hopper Superchip) \\
    GPU architecture                & Hopper, compute capability $9.0$ (sm\_$90$) \\
    GPU memory                      & $96$\,GB HBM3, $\approx 3.0$\,TB/s bandwidth \\
    Tensor Core BF16/FP16 (dense)   & $989$ TFLOPS \\
    Tensor Core FP8 (dense)         & $1979$ TFLOPS \\
    Tensor Core FP64 (dense)        & $67$ TFLOPS \\
    GPU TDP                         & $700$\,W (observed cap; configurable $450$--$1000$\,W) \\
    Host CPU                        & NVIDIA Grace, $72$ Arm Neoverse-V2 cores (aarch$64$) \\
    Host memory                     & $480$\,GB LPDDR5X \\
    CPU--GPU interconnect           & NVLink-C2C, $900$\,GB/s coherent \\
    Driver / CUDA toolkit           & NVIDIA $570.195$, CUDA $12.8$ \\
    Operating system                & Ubuntu $22.04$, Linux $6.8$ (aarch$64$) \\
    Number of GPUs / nodes          & $1$ GPU, $1$ node (no tensor or data parallelism) \\
    \bottomrule
  \end{tabular}

  \label{tab:hardware}
\end{table}

\section{Models and datasets}\label{sec:app_model_data}

All models and datasets used in this work are publicly available; we list
their canonical sources and licenses below. We did not retrain any base model;
we consume each one only as a frozen backbone for LoRA. Training and
evaluation data are likewise consumed as released, except that for the code
task, we filter the upstream Code-Feedback corpus to the Python subset (this
filtering is provided by the PiSSA repackaging that we follow). The same
sub-sampling protocol of 100\,k examples, one epoch, used by
\cite{meng2024pissa}, is applied identically across our {\fullLoRA},
{\name}, and other runs.

\paragraph{Base models}
Table~\ref{tab:models} lists the six pretrained checkpoints we fine-tune.
Access to the Llama-2 and Gemma weights is gated on Hugging Face and requires
the user to accept the corresponding license on the model card before
download; the remaining four are open-access. We use bf16 weights throughout.

\begin{table}[ht]
  \centering\small
\caption{Pretrained backbones. ``Gated'' means a one-time license click-through
  on Hugging Face is required before \texttt{huggingface-cli download} succeeds.}
  \setlength{\tabcolsep}{4pt}
  \small
  \begin{tabular}{lp{5cm}l}
    \toprule
    \textbf{Model}              & \textbf{Hugging Face identifier}                                       & \textbf{License} \\
    \hline
    LLaMA-2-7B         & \texttt{meta-llama/Llama-2-7b-hf}                             & Llama-2 Community License (gated) \\
    LLaMA-2-13B        & \texttt{meta-llama/Llama-2-13b-hf}                            & Llama-2 Community License (gated) \\
    Mistral-7B-v0.1    & \texttt{mistralai/Mistral-7B-v0.1}                            & Apache-2.0 \\
    Gemma-7B           & \texttt{google/gemma-7b}                                      & Gemma Terms of Use (gated) \\
    TinyLlama-1.1B     & \texttt{\seqsplit{TinyLlama/TinyLlama-1.1B-intermediate-step-1431k-3T}} & Apache-2.0 \\
    DeBERTa-v3-base    & \texttt{microsoft/deberta-v3-base}                            & MIT \\
    \bottomrule
  \end{tabular}
  \label{tab:models}
\end{table}

\paragraph{Training data}
The three-generation tasks follow the PiSSA training-set repackaging
\cite{meng2024pissa}, available as
\texttt{fxmeng/pissa-dataset} on Hugging Face under Apache-2.0; that bundle in
turn redistributes the upstream sources listed in Table~\ref{tab:data-train}
in a unified Alpaca-style schema. The natural-language understanding experiments use the GLUE benchmark loaded directly via
\texttt{datasets.load\_dataset("glue", task)} for each of the eight tasks
(\textsc{cola}, \textsc{sst2}, \textsc{mrpc}, \textsc{stsb}, \textsc{rte},
\textsc{qnli}, \textsc{mnli}, \textsc{qqp}); we exclude \textsc{wnli} per
common practice.

\begin{table}[ht]
  \centering\small
    \caption{Training corpora. The MetaMathQA / Code-Feedback / WizardLM rows are
  redistributed inside \texttt{fxmeng/pissa-dataset} (Apache-2.0 wrapper) in the form we actually load. We use the 143\,k WizardLM subset that is
  released directly under MIT; we do not merge into the larger ShareGPT
  augmentation that the original card describes, so the additional ShareGPT licensing terms do not apply to our experiments. GLUE itself carries the
  HF \texttt{other} tag and defers to each subtask's upstream license; we use
  GLUE only for non-commercial research.}
  \begin{tabular}{lp{3cm}p{4cm}p{3cm}}
    \toprule
    \textbf{Task}               & \textbf{Source corpus}                       & \textbf{Hugging Face id}                                  & \textbf{License} \\
    \midrule
    Math               & MetaMathQA (395\,k) & \texttt{meta-math/MetaMathQA}              & MIT \\
    Code               & Code-Feedback (Python-filtered, 105\,k)  & \texttt{\seqsplit{m-a-p/CodeFeedback-Filtered-Instruction}} & Apache-2.0 \\
    Conversation       & WizardLM-Evol-Instruct-V2 (143\,k)  & \texttt{\seqsplit{WizardLMTeam/WizardLM\_evol\_instruct\_V2\_196k}} & MIT \\
    GLUE (8 tasks)     & GLUE benchmark                      & \texttt{nyu-mll/glue}                                & \texttt{other} (per-task; see source) \\
    \bottomrule
  \end{tabular}
  \label{tab:data-train}
\end{table}

\paragraph{Evaluation data}
Table~\ref{tab:data-eval} lists the held-out evaluation sets. HumanEval+ and
MBPP+ are obtained through the \texttt{evalplus 0.3.1} package, which
augments the original HumanEval and MBPP test cases. MT-Bench prompts are
released as part of LMSYS' FastChat repository under Apache-2.0 (specifically
\texttt{fastchat/llm\_judge/data/mt\_bench/question.jsonl}); we judge
generations with GPT-4o through the OpenAI API (no model weights are
downloaded for the judge) and use the FastChat single-turn protocol, so reproducing this part of the evaluation requires an OpenAI API key in addition to the artifacts listed below.

\begin{table}[ht]
  \centering\small
\caption{Evaluation benchmarks. All test sets are loaded as released; we do
  not redistribute them.}
  \begin{tabular}{llp{5cm}}
    \toprule
    \textbf{Benchmark}          & \textbf{Hugging Face id / package}        & \textbf{License} \\
    \midrule
    GSM8K              & \texttt{openai/gsm8k}            & MIT \\
    MATH               & \texttt{hendrycks/competition\_math} & MIT \\
    HumanEval (base)   & \texttt{openai/openai\_humaneval}  & MIT \\
    HumanEval+, MBPP+  & \texttt{evalplus} v0.3.1 (PyPI)  & Apache-2.0 \\
    MBPP (base)        & \texttt{google-research-datasets/mbpp} & CC-BY-4.0 \\
    MT-Bench           & FastChat \texttt{llm\_judge} data & Apache-2.0 (questions); GPT-4o judge subject to OpenAI ToS \\
    \bottomrule
  \end{tabular}

  \label{tab:data-eval}
\end{table}

\paragraph{Compliance and intended use}
All artifacts are used in a manner consistent with their licenses. The
Llama-2 and Gemma experiments fall within the research-and-internal-use
clauses of, respectively, the Llama 2 Community License Agreement and the
Gemma Terms of Use (both gated on Hugging Face and accepted before download).
GLUE is used only for non-commercial research, deferring to each subtask's
upstream license. The remaining training and evaluation artifacts are
released under permissive open-source licenses (Apache-2.0, MIT, or
CC-BY-4.0), which we comply with by retaining each upstream's attribution
and license notice in the released code. We release no derived base-model weights; only the LoRA adapter checkpoints (which are increments to a frozen
base) and our condition-number selection regexes accompany this paper, and
each adapter inherits the license of the base it was trained on (for example, a LoRA adapter trained on Llama-2-7B remains subject to the Llama 2
Community License).


\section{Additional experiments}

\subsection{{\name} is a scalable learner}

As shown in Table~\ref{tab:supp_code_results}. We further validate the effectiveness of {\name} across models of varying scales. All cells were trained for 1 epoch on 100k examples of Code-Feedback at LoRA rank r=128. As shown, our method remains effective across different model sizes. Notably, for the smaller TinyLlama-1.1B, the frozen backbone incurs lower forward and backward computation cost, allowing our method to deliver even greater wall-clock savings. It also perform well on LLaMA-13B model and save 15.7\% training time. At the same time, it achieves performance on par with {\fullLoRA}, and in several cases even surpasses it. 

\begin{table}[t]
\centering
\caption{Code-Feedback fine-tuning evaluated on HumanEval / HumanEval+ and MBPP / MBPP+ (pass@1, evalplus 0.3.1).}
\label{tab:supp_code_results}
\resizebox{0.8\linewidth}{!}{
\begin{tabular}{cccccccc}
\toprule
\multirow{2}{*}{\textbf{Model}} & \multirow{2}{*}{\textbf{Method}} & \multirow{2}{*}{\textbf{Time(↓)}} & \multicolumn{2}{c}{\textbf{HumanEval}} & \multicolumn{2}{c}{\textbf{MBPP}} \\
\cmidrule(lr){4-5} \cmidrule(lr){6-7}
 & & & \textbf{base(↑)} & \textbf{plus(↑)} & \textbf{base(↑)} & \textbf{plus(↑)} \\
\midrule
\multirow{2}{*}{\texttt{TinyLlama-1.1B}}
            & {\fullLoRA} & 2369s          & \textbf{13.40} & 11.00 & \textbf{27.20} & \textbf{21.20} \\
            & {\name}     & \textbf{1773s(-25.2\%)} & \textbf{13.40} & \textbf{12.20} & 24.60 & 19.00 \\
\midrule
\multirow{2}{*}{\texttt{LLaMA 2-7B}}
            & {\fullLoRA} & 7221s          & \textbf{21.80} & \textbf{19.50} & 33.40 & 27.80 \\
            & {\name}     & \textbf{6258s(-13.3\%)} & 21.30 & 18.90 & \textbf{37.60} & \textbf{29.60} \\
\midrule
\multirow{2}{*}{\texttt{LLaMA 2-13B}}
            & {\fullLoRA} & 13048s         & \textbf{31.10} & \textbf{28.00} & 46.30 & 39.40 \\
            & {\name}     & \textbf{11001s(-15.7\%)} & 30.50 & 27.40 & \textbf{47.10} & \textbf{39.70} \\
\bottomrule
\end{tabular}
}
\end{table}

\end{document}